\documentclass[11pt,a4paper]{article}
\usepackage[hyperref]{acl2017}
\usepackage{times}
\usepackage{latexsym}

\usepackage{url}

\aclfinalcopy % Uncomment this line for the final submission

\usepackage{amssymb}
\usepackage{multicol}
\usepackage{amsmath}
\usepackage{graphicx}
\definecolor{green4}{rgb/hsb}{0.0,0.5,0.0/0.3,1.0,0.5}

\newcommand{\mycomment}[1]{}
\newcommand{\ignore}[1]{}

\DeclareMathOperator*{\sig}{\sigma}

\newcommand{\entail}{{\Rightarrow}}
\newcommand{\notentail}{{\,/\!\!\!\!{\Rightarrow}}}

\newcommand{\bie}{{\raisebox{1.8pt}{\makebox[0pt][l]{\hspace{1.3pt}${\scriptscriptstyle >}$}${\scriptscriptstyle \bigcirc}$}}}

\title{ Learning Word Embeddings for Hyponymy with  \\
  Entailment-Based Distributional Semantics \\
}

\author{  James Henderson
  \thanks{~ This paper reports work done while the author was at 
    Xerox Research Centre Europe.} \\
  Idiap Research Institute \\
  {\tt james.henderson@idiap.ch} \\
}

\date{}

\begin{document}
\maketitle

\begin{abstract}
Lexical entailment, such as hyponymy, is a fundamental issue in the semantics
of natural language.  This paper proposes distributional semantic models which
efficiently learn word embeddings for entailment, using a recently-proposed
framework for modelling entailment in a vector-space.  These models postulate
a latent vector for a pseudo-phrase containing two neighbouring word vectors.
We investigate both modelling words as the evidence they contribute about this
phrase vector, or as the posterior distribution of a one-word phrase vector,
and find that the posterior vectors perform better.  The resulting word
embeddings outperform the best previous results on predicting hyponymy between
words, in unsupervised and semi-supervised experiments.
\end{abstract}

\section{Introduction}

Modelling entailment, such as hyponymy, is a fundamental issue in the
semantics of natural language, and there has been a lot of interest in
modelling entailment using vector-space representations, particularly for
lexical entailment relations such as hyponymy.  Entailment is the relation of
information inclusion, meaning that $y$ entails $x$ if and only if everything
that is known given $x$ is also known given $y$.  As such, representations
which support entailment need to encode what is known, versus what is unknown.

Although much work has used vector-space embeddings of words in models of
entailment, few models have developed vector-space embeddings which
intrinsically model entailment.  The exceptions have been \citet{Vilnis15},
who use variances to represent the amount of information about a continuous
space, and \citet{Henderson16_acl}, who use probabilities to represent the
amount of information about a discrete space.  In this work we use the
framework from \citet{Henderson16_acl} to develop new distributional semantic
models of entailment between words.

In the framework of \citet{Henderson16_acl}, each dimension of the
vector-space represents something that might be known, and continuous vectors
represent probabilities of these features being known or unknown.
\citet{Henderson16_acl} illustrate their framework by proposing a
reinterpretation of existing Word2Vec \citep{word2vec1} word embeddings, which
successfully predicts hyponymy with an unsupervised model.  To motivate this
reinterpretation of existing word embeddings, they propose a model of
distributional semantics and argue that, under this reinterpretation, the
Word2Vec training objective approximates the training objective of this
distributional semantic model.

In this paper, we implement this distributional semantic model and train new
word embeddings using the exact objective.  This results in embeddings which
directly encode what is known and unknown given a word, thus not requiring any
reinterpretation to predict hyponymy.  The distributional semantic model
postulates a latent pseudo-phrase vector for the unified semantics of a word
and its neighbouring context word.  This latent vector must entail the
features in both words' vectors and must be consistent with a prior over
semantic vectors, thereby modelling the redundancy and consistency between the
semantics of two neighbouring words.

Our analysis of this entailment-based model of a word in context leads us to
hypothesise that the word embeddings suggested by
\linebreak\citet{Henderson16_acl} are in fact not the best way to extract
information about the semantics of a word from this model.  They propose using
a vector which represents the evidence about known features given the word.
We propose to instead use a vector which represents the posterior distribution
of known features for a phrase containing only the word.  This posterior
vector includes both the evidence from the word and its indirect consequences
via the constraints imposed by the prior.  Our efficient implementation of
this model allows us to test this hypothesis by outputting either the evidence
vectors or the posterior vectors as word embeddings.

To evaluate these word embeddings, we predict hyponymy between words, in both
an unsupervised and semi-supervised setting.  Given the word embeddings for
two words, we measure whether they are a hypernym-hyponym pair using an
entailment operator from \citep{Henderson16_acl} applied to the two
embeddings.  We find that using the evidence vectors performs as well as
reinterpreting Word2Vec embeddings, confirming the claims of equivalence by
\citet{Henderson16_acl}.  But we also find that using the posterior vectors
performs significantly better, confirming our hypothesis that posterior
vectors are better, and achieving the best published results on this benchmark
dataset.  In addition to these unsupervised experiments, we evaluate in a
semi-supervised setting and find a similar pattern of results, again achieving
state-of-the-art performance.

In the rest of this paper, section~\ref{sec:sem} presents the formal
framework we use for modelling entailment in a vector space, the
distributional semantic models, and how these are used to predict hyponymy.
Section~\ref{sec:related} discusses additional related work, and then
section~\ref{sec:results} presents the empirical evaluation on hyponymy
detection, in both unsupervised and semi-supervised experiments.  Some
additional analysis of the induced vectors is presented in
section~\ref{sec:discussion}.

\section{Distributional Semantic Entailment}
\label{sec:sem}

Distributional semantics uses the distribution of contexts in which a word
occurs to induce the semantics of the word
\citep{Harris54,Deerwester1990,Schutze93}.  The Word2Vec model
\citep{word2vec1} introduced a set of refinements and computational
optimisations of this idea which allowed the learning of vector-space
embeddings for words from very large corpora with very good semantic
generalisation.  \citet{Henderson16_acl} motivate their reinterpretation the
Word2Vec Skipgram \citep{word2vec1} distributional semantic model with an
entailment-based model of the semantic relationship between a word and its
context words.  We start by explaining our interpretation of the
distributional semantic model proposed by \citet{Henderson16_acl}, and then
propose our alternative models.

\begin{figure*}
  \begin{center}
    \includegraphics[scale=0.7]{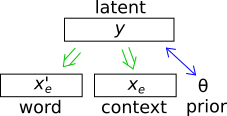}
    ~~~    ~~~    ~~~    
    \includegraphics[scale=0.7]{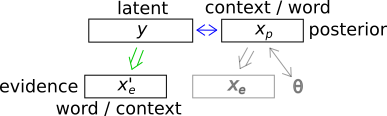}
    %\vspace{-4ex}
  \end{center}
  \caption{The distributional semantic model of a word and its context (left),
  and its approximation in the word2hyp models (right).}
  \label{fig:distsem}
\end{figure*}

\citet{Henderson16_acl} postulate a latent vector $y$ which is the consistent
unification of the features of the middle word $x^\prime_e$ and the neighbouring
context word $x_e$, illustrated on the left in
figure~\ref{fig:distsem}.\footnote{Note that ``$x_e$'' is being used here as
  the name of a whole vector, not to be confused with ``$x_i$'', which refers
  to element $i$ in vector $x$.}  We can think of the latent vector $y$ as
representing the semantics of a pseudo-phrase consisting of the two words.
The unification requirement is defined as requiring that $y$ entail both
words, written $y\entail x^\prime_e$ and $y\entail x_e$.  The consistency
requirement is defined as $y$ satisfying the constraints imposed by a prior
$\theta(y)$.  This approach models the relationship between the semantics of a
word and its context as being redundant and consistent.  If $x^\prime_e$ and
$x_e$ share features, then it will be easier for $y$ to satisfy both
$y\entail x^\prime_e$ and $y\entail x_e$.  If the features of $x^\prime_e$ and
$x_e$ are consistent, then it will be easier for $y$ to satisfy the
prior $\theta(y)$.

\subsection{The Reinterpretation of Word2Vec}

\citet{Henderson16_acl} formalise the above model using their
entailment-vectors framework.  This framework models distributions over
discrete vectors where a one in position $i$ means feature $i$ is known and a
zero means it is unknown.  Entailment $y\entail x$ requires that the ones in
$x$ are a subset of the ones in $y$, so $1\entail 1$, $0\entail 0$ and
$1\entail 0$, but $0\notentail 1$.  Distributions over these discrete vectors
are represented as continuous vectors of log-odds $X$, so
$P(x_i{=}1)=\sig(X_i)$, where $\sig$ is the logistic sigmoid.  The probability
of entailment $y\entail x$ between two such ``entailment vectors'' $Y,X$ can
be measured using the operator $\bie$:\footnote{We use lowercase variables
  $x,y$ to refer to discrete vectors and uppercase variables $X,Y$ to refer to
  their associated entailment vectors.}
\begin{align}
  \log &P(y\entail x ~|~ Y,X) ~\approx~
  \nonumber
  \\
  &
  Y \bie X
  ~\equiv~ \sig({-}Y) \cdot \log\sig({-}X)
  \label{eqn:bie}
  %\\[-3.7ex]\nonumber
\end{align}
For each feature $i$ in the vector, it calculates the expectation according to
$P(y_i)$ that, either $y_i{=}1$ and thus the log-probability is zero, or
$y_i{=}0$ and thus the log-probability is $\log P(x_i{=}0)$ (given that
$\sig({-}X_i)=(1-\sig(X_i))\approx P(x_i{=}0)$).

\begin{figure}[htb]
  %\vspace{-0.1cm}
  \centerline{\includegraphics[width=0.9\linewidth]{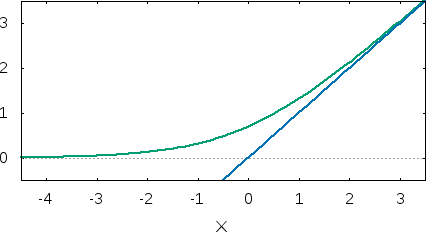}}
  \vspace{-0.4cm}
  \caption{ The function {\color{green4} ${-}\log\sig({-}X)$} used in\linebreak inference and the
    $\bie$ operator, versus {\color{blue} $X$}. 
  }
  \label{fig:lsc}
\end{figure}

\citet{Henderson16_acl} formalise the model on the left in
figure~\ref{fig:distsem} by first inferring the optimal latent vector
distribution $Y$ (equation (\ref{eqn:inf1})), and then scoring how well the
entailment and prior constraints have been satisfied (equation
(\ref{eqn:sim1})).
\begin{align}
  \max_Y&(E_{Y,X^\prime_e,X_e} \log P(y\entail x^\prime_e,\,y\entail x_e,\,y) )
  \nonumber
  \\
  &\!\!\!\!\!\!\approx
  Y\bie X^\prime_e + Y\bie X_e + ({-}\sig({-}Y))\cdot \theta(Y)
  \label{eqn:sim1}
  \\
  \mbox{where}\hspace{-2ex}&
  \nonumber
  \\&\!\!\!\! %\hspace{2ex}
  Y =  - \log\sig({-}X^\prime_e) + - \log\sig({-}X_e) + \theta(Y) 
  \label{eqn:inf1}
\end{align}
where $E_{Y,X^\prime_e,X_e}$ is the expectation over the distribution defined
by the log-odds vectors $Y,X^\prime_e,X_e$, and $\log$ and $\sig$ are applied
componentwise.  The term $\theta(Y)$ is used to indicate the net effect of the
prior on the vector $Y$.  Note that, in the formula (\ref{eqn:inf1}) for
inferring $Y$, the contribution ${-}\log\sig({-}X)$ of each word vector is
also a component of the definition of $Y\bie X $ from
equation~(\ref{eqn:bie}).
In this way, the score for measuring how well the entailment has been
satisfied is using the same approximation as used in the inference to satisfy
the entailment constraint.  This function ${-}\log\sig({-}X)$ is a non-negative
transform of $X$, as shown in figure~\ref{fig:lsc}.  Intuitively, for an
entailed vector $x$, we only care about the probability that $x_i{=}1$
(positive log-odds $X_i$), because that constrains the entailing vector $y$ to
have $y_i{=}1$ (adding to the log-odds $Y_i$).

The above model cannot be mapped directly to the Word2Vec model because
Word2Vec has no way to model the prior $\theta(Y)$.  On the other hand, the
Word2Vec model postulates two vectors for every word, compared to one in the
above model.  \citet{Henderson16_acl} propose an approximation to the above
model which incorporates the prior into one of the two vectors, resulting in
each word having one vector $X_e$ as above plus another vector $X_p$
with the prior incorporated.
\begin{align}
  X_p &\approx  {-}\log\sig({-}X_e) + \theta(Y) 
  \label{eqn:xp}
\end{align}
Both vectors $X_e$ and $X_p$ are parameters of the model, which need to be
learned.  Thus, there is no need to explicitly model the prior, thereby
avoiding the need to choose a particular form for the prior $\theta$, which in
general may be very complex.

This gives us the following score for how well the constraints of this model
can be satisfied.
\begin{align}
  \max_Y&(E_{Y,X^\prime_e,X_p} \log P(y\entail x^\prime_e,\,y\entail x_e,\,y) )
  \nonumber
  \\
  &\approx
    %{\simm}(X^\prime_e,X_p) \equiv
    Y\bie X^\prime_e + ({-}\sig({-}Y))\cdot X_p
  \label{eqn:sim2}
  \\
  \mbox{where}\hspace{-2ex}&
  \nonumber
  \\& %\hspace{2ex}
  Y =  - \log\sig({-}X^\prime_e) + X_p
  \label{eqn:inf2}
\end{align}

In \citep{Henderson16_acl}, score~(\ref{eqn:sim2}) is only used to provide a
reinterpretation of Word2Vec word embeddings.  They show that a transformation
of the vectors output by Word2Vec (``W2V {\em u.d.$\bie$}'' below) can be seen
as an approximation to the evidence vector $X_e$.  In
Section~\ref{sec:results}, we empirically test this hypothesis by directly
training $X_e$ (``W2H evidence'' below) and comparing the results to those
with reinterpreted Word2Vec vectors.

\subsection{New Distributional Semantic Models}

In this paper, we implement distributional semantic models based on
score~(\ref{eqn:sim2}) and use them to train new word embeddings.
We call these models the
Word2Hyp models, because they are based on Word2Vec but are designed to
predict hyponymy.  

To motivate our models, we provide a better understanding of the model behind
score~(\ref{eqn:sim2}).  In particular, we note that although we want $X_p$ to
approximate the effects $\theta(Y)$ of the prior as in equation~\ref{eqn:xp},
in fact $X_p$ is only dependent on one of the two words, and thus can only
incorporate the portion of $\theta(Y)$ which arises from that one word.
Thus, a better understanding of $X_p$ is provided by
equation~(\ref{eqn:post}).
\begin{align}
  X_p
  %&\approx  {-}\log\sig({-}X_e) + \theta(Y) 
  %\nonumber
  %\\
  &\approx  {-}\log\sig({-}X_e) + \theta(X_p) 
  \label{eqn:post}
\end{align}
In this framework, equation~(\ref{eqn:post}) is exactly the same formula as
would be used to infer the vector for a single-word phrase (analogously to
equation~(\ref{eqn:inf1})).

This interpretation of the approximate model in equation~\ref{eqn:sim2} is
given on the right side of figure~\ref{fig:distsem}.  As shown, $X_p$ is
interpreted as the {\em posterior} vector for a single-word phrase, which
incorporates the evidence and the prior for that word.  In contrast,
$X^\prime_e$ is just the {\em evidence} about $Y$ provided by the other word.
This model, as argued above, approximates the model on the left side in
Figure~\ref{fig:distsem}.  But the grey part of the figure does not need to be
explicitly modelled.

This interpretation suggests that the posterior vector $X_p$ should be a better
reflection of the semantics of the word than the evidence vector $X_e$, since
it includes both the direct evidence for some features and their indirect
consequences for other features.  We test this hypothesis empirically in
Section~\ref{sec:results}.

To implement our distributional semantic models, we define new versions of the
Word2Vec code \citep{word2vec1,word2vec2_nips}.  The Word2Vec code trains two
vectors for each word, where negative sampling is applied to one of these
vectors, and the other is the output vector.  This applies to both the
Skipgram and CBOW versions of training.  Both versions also use a dot product
between vectors to try to predict whether the example is a positive or
negative sample.
We simply replace this dot product with score~(\ref{eqn:sim2}) directly in the
Word2Vec code, leaving the rest of the algorithm unchanged.  We make this
change in one of two ways, one where the output vector corresponds to the
evidence vector $X_e$, and one where the output vector corresponds to the
posterior vector $X_p$.
We will refer to the model where $X_p$ is output as the ``posterior'' model,
and the model where $X_e$ is output as the ``evidence'' model.
Both these methods can be applied to both the Skipgram and CBOW models, giving
us four different models to evaluate.

\subsection{Modelling Hyponymy}

The proposed distributional semantic models output a word
embedding vector for every word in the vocabulary, which are directly
interpretable as entailment vectors in the entailment framework.  Thus, to
predict lexical entailment between two words, we can simply apply the $\bie$
operator to their vectors, to get an approximation of the log-probability of
entailment.  

We evaluate these entailment predictions on hyponymy detection.
Hyponym-hypernym pairs should have associate embeddings $Y,X$ which have a
higher entailment scores $Y\bie X$ than other pairs.  We rank the word pairs
by the entailment scores for their embeddings, and evaluate this ranked list
against the gold hyponymy annotations.
We evaluate on hyponymy detection because it reflects a direct form of
lexical entailment;
the semantic features of a hypernym (e.g.\ ``animal'') should be included in
the semantic features of the hyponym (e.g.\ ``cat'').  Other forms of lexical
entailment would benefit from some kind of reasoning or world knowledge, which
we leave to future work on compositional models.

\section{Related Work}
\label{sec:related}

In this paper we propose a distributional semantic model which is based on
entailment.  Most of the work on modelling entailment with vector space
embeddings has simply used distributional semantic vectors within a model of
entailment, and is therefore not directly relevant here.  See
\citep{Shwartz17} for a comprehensive review of such measures.
\citet{Shwartz17} evaluate these measure as unsupervised models of hyponymy
detection and run experiments on a number of hyponymy datasets.
We report their best comparable result in Table~\ref{tab:results-unsup}.

\citet{Vilnis15} propose an unsupervised model of entailment in a vector
space, and evaluate it on hyponymy detection.  Instead of representing words
as a point in a vector 
space, they represent words as a Gaussian distribution over points in a vector
space.  The variance of this distribution in a given dimension indicates the
extent to which the dimension's feature is unknown, so they use KL-divergence
to detect hyponymy relations.
Although this model has a nice theoretical motivation, the word
representations are more complex and training appears to be more
computationally expensive than the method proposed here.

The semi-supervised model of \citet{Kruszewski15}
learns a discrete Boolean vector space for predicting hyponymy.  But they
do not propose any unsupervised method for learning these vectors.

\citet{weeds2014learning} report hyponymy detection results for a number of
unsupervised and semi-supervised models.
They propose a semi-supervised evaluation methodology where the words in the
training and test sets are disjoint, so that the supervised component must
learn about the unsupervised vector space and not about the individual words.
Following \citet{Henderson16_acl}, we replicate their experimental setup in
our evaluations, for both unsupervised and semi-supervised models, and compare
to the best results among the models evaluated by \citet{weeds2014learning},
\citet{Shwartz17} and \citet{Henderson16_acl}.

\section{Evaluation of Word Embeddings}
\label{sec:results}

We evaluate on hyponymy detection in both a fully unsupervised setup and a
semi-supervised setup. 
In the semi-supervised setup, we use labelled hyponymy data to train a linear
mapping from the unsupervised vector space to a new vector space with the
objective of correctly predicting hyponymy relations in the new vector space.
This prediction is done with the same (or equivalent) entailment operator
as for the unsupervised experiments (called ``map $\bie$'' in
Table~\ref{tab:results-semi}).

We replicate the experimental setup of \citet{weeds2014learning},
using their selection of hyponym-hypernym pairs
from
the BLESS dataset \citep{bless-dataset}, which consists of noun-noun pairs,
including 50\% positive hyponymy pairs plus 50\% negative pairs consisting of
some other hyponymy pairs reversed, some pairs in other semantic relations,
and some random pairs.
As in \citep{weeds2014learning}, our semi-supervised experiments use ten-fold
cross validation, where each fold has items removed from the training set if
they contain a word that also occurs in the testing set.

The word embedding vectors which we train have 200 dimensions and were trained
using our Word2Hyp modification of the Word2Vec code (with default
settings), trained on a corpus of half a billion words of Wikipedia.
We also replicate the
approach of \citet{Henderson16_acl} by training Word2Vec embeddings on this data.

To quantify performance on hyponymy detection, for each model
we rank the list of pairs according to the score given by the model, and
report two measures of performance for this ranked lists.
The ``{\em 50\% Acc}\/'' measure treats the first half of the list as labelled
positive and the second half as labelled negative.  This is motivated by the
fact that we know a~priori that the proportion of positive examples has been
artificially set to (approximately) 50\%.
Average precision is a measure of the accuracy for ranked lists, used in
Information Retrieval and advocated as a measure of hyponymy detection by
\citet{Vilnis15}.  For each positive example, precision is measured at the
threshold just below that example, and these precision scores are averaged
over positive examples.  For cross validation, we average over the union of
positive examples in all the test sets.  Both these measures are reported
(when available) in Tables~\ref{tab:results-unsup}
and ~\ref{tab:results-semi}.

\subsection{Unsupervised Hyponymy Detection} % and Direction Classification
\label{sec:unsuper}

The first set of experiments evaluate the different embeddings in their
unsupervised models of hyponymy detection.  Results are shown in  Table~\ref{tab:results-unsup}.
Our principal point of comparison is the best results from \citep{Henderson16_acl}
(called ``W2V GoogleNews'' in
Table~\ref{tab:results-unsup}).  They use the pre-existing publicly available GoogleNews
word embeddings, which were
trained with the Word2Vec software on 100 billion words of the GoogleNews
dataset, and have 300 dimensions.  To provide a more direct comparison, we
replicate the model of \citet{Henderson16_acl} but using the same embedding
training setup as for our Word2Hyp model (``W2V Skip'').
Both cases use their proposed reinterpretation of these
vectors for predicting entailment (``{\em u.d.}$\bie$'').
We also report the best results from
\citet{weeds2014learning} and the best results from \citep{Shwartz17}.
For our proposed Word2Hyp distributional semantic models (``W2H''), we report results
for the four combinations of using the CBOW or Skipgram (``Skip'') model to train the
evidence or posterior vectors.

\begin{table}[tb]
\begin{center}
\begin{tabular}{|@{~}l@{~}c@{~}|@{~~}l@{~~}l@{~}|}
\hline
\multicolumn{1}{|c}{embeddings} &  \!\!operator %&  supervision 
&  {\small\em 50\% Acc} & {\small\em Ave Prec} \\ %&  \!\!{\em Dir Acc} \\
\hline\hline
\multicolumn{2}{|c|@{~~}}{Weeds et.al., 2014}       & 	58\% & 	~-- \\ %& 	~-- \\
\multicolumn{2}{|c|@{~~}}{Shwartz et.al., 2017}       & 	~-- & 	44.1\% \\ %& 	~-- \\
\hline
W2V GoogleNews & {\em u.d.}$\bie$ & 	64.5\%* & 	~-- \\ %& 	68.8\% \\ 	% 68.8\%
\hline
W2V {\sc cbow} & {\em u.d.}  $\bie$ & 	53.2\% & 	55.2\% \\ %& 	49.4\% \\
W2H Skip \hfill evidence &  $\bie$ & 	59.5\% & 	57.8\% \\ %& 	64.5\% \\
W2H {\sc cbow} \hfill evidence &  $\bie$ & 	61.8\% & 	66.4\% \\ %& 	65.3\% \\
W2V Skip & {\em u.d.}$\bie$ & 	62.1\% & 	67.6\% \\ %& 	68.6\% \\
W2H {\sc cbow} \hfill posterior &  $\bie$ & 	68.1\%* & 	{\bf 70.8}\% \\ %& 	77.3\% \\
W2H Skip \hfill posterior &  $\bie$ & 	{\bf 69.6}\% & 	68.9\% \\ %& 	{\bf 78.8}\% \\
\hline
\end{tabular}
\end{center}
\caption{ Hyponymy detection accuracies
  %on the BLESS data from \citep{weeds2014learning}, 
  %using the Google-News word embeddings
  %(except {\em map-wiki} $\bie$ which uses Wikipedia word embeddings), 
  ({\em 50\%  Acc}) and average precision ({\em Ave Prec}), 
  %and hyponymy direction classification ({\em Dir Acc}),
  in the unsupervised experiments.  
  For the accuracies, * marks a significant improvement.
  %{\em 50\% Acc} classification thresholds are set to output 50\% positives.
  %{\em Weeds et.al.} uses different embeddings and a classifier instead of a
  %vector mapping. 
  %\vspace{-2ex}
}
\label{tab:results-unsup}
\end{table}

The best unsupervised model of \citet{weeds2014learning}
and the two Word2Hyp models with evidence vectors perform similarly.  The
reinterpretation of Word2Vec vectors (``W2V GoogleNews {\em u.d.}$\bie$'')
performs better, but when the same method is applied to the smaller Wikipedia
corpus (``W2V Skip {\em u.d.}$\bie$''), this difference all but disappears.
This confirms the hypothesis of \citet{Henderson16_acl} that the
reinterpretation of Word2Vec vectors and the evidence vectors from
Word2Hyp are approximately equivalent.

However, even with this smaller corpus, using the proposed posterior vectors
from the Word2Hyp model are significantly more accurate than the
reinterpretation of Word2Vec vectors.  This confirms the hypothesis that the
posterior vectors from the Word2Hyp model are a better model of the semantics
of a word than the evidence vectors suggested by \citet{Henderson16_acl}.

Using the CBOW model or the Skipgram model makes only a small difference.
The average precision score shows the same pattern as the accuracy.

To allow a direct comparison to the model of \citet{Vilnis15}, we also
evaluated the unsupervised models on the hyponymy data from \citep{Baroni12},
which is not as carefully designed to evaluate hyponymy as the
\citep{weeds2014learning} data.
Both the evidence and posterior vectors of the Word2Hyp CBOW model achieved 
average precision (81\%, 80\%) which is not significantly different from the
best model of \citet{Vilnis15} (80\%).

\begin{table}[tb]
\begin{center}
\begin{tabular}{|@{~}l@{~}c@{~}|@{~~}l@{~~}l@{~}|}
\hline
\multicolumn{1}{|c}{embeddings} &  \!\!operator %&  supervision 
&  {\small\em 50\% Acc} & {\small\em Ave Prec} \\ %&  \!\!{\em Dir Acc} \\
\hline\hline
\multicolumn{2}{|c|@{~~}}{Weeds et.al., 2014}        	& 	75\% & 	~-- \\ %& 	~-- \\
\hline
W2V GoogleNews & {\em map} $\bie$          & 	80.1\% & 	~-- \\ %&   	90.0\% \\ 	% 86.3\%
\hline
W2V Skip   & {\em map} $\bie$        &  	81.9\% &  	88.3\% \\ %& 	90.8\% \\
W2H {\sc cbow} \hfill evidence & {\em map} $\bie$ & 	83.3\% & 	90.3\% \\ %& 	{\bf 95.9}\% \\
W2V {\sc cbow}          & {\em map} $\bie$     &     	84.6\% &  	91.5\% \\ %&	95.2\% \\
W2H Skip \hfill evidence & {\em map} $\bie$ & 	84.8\% & 	90.9\% \\ %& 	93.5\% \\
W2H Skip \hfill posterior & {\em map} $\bie$ & 	85.5\% & 	91.3\% \\ %& 	94.3\% \\
W2H {\sc cbow} \hfill posterior & {\em map} $\bie$ & 	{\bf 86.0}\% & 	{\bf 92.8}\% \\ %& 	{\bf 95.9}\% \\
\hline
\end{tabular}
\end{center}
\caption{ Hyponymy detection accuracies
  %on the BLESS data from \citep{weeds2014learning}, 
  %using the Google-News word embeddings
  %(except {\em map-wiki} $\bie$ which uses Wikipedia word embeddings), 
  ({\em 50\%  Acc}) and average precision ({\em Ave Prec}), 
  %and hyponymy direction classification ({\em Dir Acc}),
  in the semi-supervised experiments.  
  %{\em 50\% Acc} classification thresholds are set to output 50\% positives.
  %{\em Weeds et.al.} uses different embeddings and a classifier instead of a
  %vector mapping. 
  %\vspace{-2ex}
}
\label{tab:results-semi}
\end{table}

\subsection{Semi-supervised Hyponymy Detection} % and Direction Classification

The semi-supervised experiments train a linear mapping from
each unsupervised vector space to a new vector space, where the entailment
operator $\bie$ is used to predict hyponymy (``map $\bie$'').

The semi-supervised results (shown in 
Table~\ref{tab:results-semi})\footnote{It is not clear how to measure
  significance for cross-validation results, so we do not attempt to do so.}
no longer show an advantage of GoogleNews vectors over Wikipedia
vectors for the reinterpretation of Word2Vec vectors.  And the advantage of
posterior vectors over the evidence vectors is less pronounced.  However, the
two posterior vectors still perform much better than all the previously
proposed models, achieving 86\% accuracy and nearly 93\% average precision.
These semi-supervised results confirm the results from the unsupervised
experiments, that Word2Vec embeddings and Word2Hyp evidence embeddings perform
similarly, but that using the posterior vectors of the Word2Hyp model perform
better.

\subsection{Training Times}

Because the similarity measure in equation~\ref{eqn:sim2} is more complex than a simple dot
product, training a new distributional semantic model is slower than with the
original Word2Vec code.  In our experiments, training took about 8 times
longer for the CBOW model and about 15 times longer for the Skipgram model.
This meant that Word2Hyp CBOW trained about 8 times faster than Word2Hyp
Skipgram.  As in the Word2Vec code, we used a quadrature approximation
(i.e.\ a look-up table) to speed up the computation of the sigmoid function,
and we added the same technique for computing the log-sigmoid function.

\subsection{Discussion}
\label{sec:discussion}

The relative success of our distributional semantic models at unsupervised
hyponymy detection indicates that they are capturing some aspects of lexical
entailment.  But the gap between the unsupervised and semi-supervised results
indicates that other features are also being captured.
This is not surprising, since many other factors influence the co-occurrence
statistics of words.

To get a better understanding of these word embeddings, we ranked them by
degree of abstractness.
Table~\ref{tab:abstract} shows the most abstract and least abstract frequent
words that occur in the hyponymy data.  To measure abstractness, we used our
best unsupervised embeddings and measured how well they are entailed by the
zero log-odds vector, which represents a uniform half probability of knowing
each feature.  For a vector to be entailed by the zero vector, it must be
that its features are mostly probably unknown.  The less you know given a
word, the more abstract it is.

\begin{table}[tb]
\begin{center}
\begin{tabular}{|ll|ll|}
\hline
\multicolumn{2}{|c|}{most abstract} & \multicolumn{2}{c|}{least abstract} \\
\hline
{\em something} & {\em necessity}	& { \dots} & {\em fork}   \\[-0.3ex]
{\em anything} & {\em sense}  	& {\em hockey} & {\em housing}  \\[-0.3ex]
{\em end} & {\em back}     	& {\em republican} & {\em elm}  \\[-0.3ex]
{\em inside} & {\em saw}    	& {\em hull} & {\em primate}  \\[-0.3ex]
{\em good} & { \dots}      	& {\em cricket} & {\em fur}  \\
\hline
\end{tabular}
\end{center}
\caption{ Ranking of the abstractness (${\bf 0}\,\bie X$) of frequent words from the
  hyponymy dataset, using Word2Hyp-Skipgram-posterior embeddings.  
}
\label{tab:abstract}
\end{table}

An initial ranking found that six of the top ten abstract words had frequency
less than 300 in the Wikipedia data, but none of the ten least abstract terms
were infrequent.  This indicates a problem with the current method, since
infrequent words are generally very specific (as was the case for
these low-frequency words, {\em submissiveness, implementer, overdraft,
  ruminant, warplane,} and {\em londoner}).  Although this is an interesting
characteristic of the method, the terms themselves seem to be noise, so we
rank only terms with frequency greater than 300.

The most abstract terms in table~\ref{tab:abstract} include some clearly
semantically abstract terms, in particular {\em something} and {\em anything}
are ranked highest.
Others may be affected by lexical ambiguity, since the model does not
disambiguate words by part-of-speech (such as {\em end}, {\em good}, {\em
  sense}, {\em back}, and {\em saw}).  The least abstract terms are mostly very
semantically specific, but it is indicative that this list includes {\em
  primate}, which is an abstract term in Zoology but presumably occurs in very
specific contexts in Wikipedia.

\section{Conclusions}

In this paper, we propose distributional semantic models for efficiently
training word embeddings which are specifically designed to capture semantic
entailment.  This work builds on the work of \citet{Henderson16_acl}, who
propose a  framework for modelling entailment in a vector-space, and a distributional
semantic model for reinterpreting Word2Vec word embeddings.  Our contribution
differs from theirs in that we train new word embeddings, and we choose
different vectors in the model to output as word embeddings.  Empirical
results on unsupervised and semi-supervised hyponymy detection confirm that
the model's evidence vectors, which \citet{Henderson16_acl} suggest to use, do
indeed perform equivalently to their reinterpretation of Word2Vec vectors.  But
these experiments also show that the model's posterior vectors, which we
propose to use, perform significantly better, outperforming all previous
results on this task.

The success of this distributional semantic model demonstrates that the
entailment vector framework can be effective at extracting information about
lexical entailment from the redundancy and consistency of words with their
contexts in large text corpora.  This result suggests future work on modelling
other indirect evidence about semantics using the entailment vector framework.

\bibliography{henderson_word2hyp_arxiv}
\bibliographystyle{acl_natbib}

\end{document}